\pdfoutput=1

\documentclass[11pt]{article}

\usepackage{ACL2023} 

\usepackage[hang,flushmargin]{footmisc}

\usepackage[nopar]{lipsum} 
\usepackage{hyperref}
\usepackage[T1]{fontenc}

\usepackage[utf8]{inputenc}

\usepackage{microtype}
\usepackage{array}
\newcolumntype{L}[1]{>{\raggedright\let\newline\\\arraybackslash\hspace{0pt}}m{#1}}
\newcolumntype{C}[1]{>{\centering\let\newline\\\arraybackslash\hspace{0pt}}m{#1}}
\newcolumntype{R}[1]{>{\raggedleft\let\newline\\\arraybackslash\hspace{0pt}}m{#1}}

\usepackage{inconsolata}
\usepackage{times}
\usepackage{latexsym}
\usepackage{graphicx}
\usepackage{listings}
\usepackage{multirow}
\usepackage{amssymb}
\usepackage{amsbsy}
\usepackage{amsmath}
\usepackage{colortbl}
\usepackage{geometry} 
\usepackage{enumitem}
\setlist{nolistsep}
\usepackage{todonotes}
\usepackage{caption,subcaption}
\usepackage{algorithmic}
\newcommand{\ModelName}{\textsc{Lyra}}

\usepackage{booktabs}

\newcommand*\samethanks[1][\value{footnote}]{\footnotemark[#1]}

\title{Unsupervised Melody-to-Lyric Generation}

\author{
    Yufei Tian\textsuperscript{\rm 1}\thanks{~~Work was done when the authors interned at Amazon. Yufei Tian is responsible for running all experiments related to this work. Anjali Narayan-Chen, Shereen Oraby, Alessandra Cervone, Gunnar Sigurdsson, Chenyang Tao, Wenbo Zhao, Jing Huang, and Nanyun Peng attended Yufei’s weekly project meeting during her summer intern and provided feedback/guidance throughout the project. Yiwen Chen helped to curate evaluation data, as well as designing and providing valuable insights to the human evaluation guidelines. Tagyoung Chung and Nanyun Peng decided on the high-level research direction before the internship started.}, Anjali Narayan-Chen\textsuperscript{\rm 2}, Shereen Oraby\textsuperscript{\rm 2}, Alessandra Cervone\textsuperscript{\rm 2}, \\
    \textbf{Gunnar Sigurdsson\textsuperscript{\rm 2}, Chenyang Tao\textsuperscript{\rm 2}, Wenbo Zhao\textsuperscript{\rm 2}, Yiwen Chen\textsuperscript{\rm 3}\samethanks{},} \\
    \textbf{Tagyoung Chung\textsuperscript{\rm 2}, Jing Huang\textsuperscript{\rm 2}, Nanyun Peng\textsuperscript{\rm 1,2}} \\
    \textsuperscript{\rm 1} University of California, Los Angeles, \textsuperscript{\rm 2} Amazon Alexa AI, \textsuperscript{\rm 3} University of Cambridge \\
    \texttt{\{yufeit,violetpeng\}@cs.ucla.edu}\\ 
    \texttt{\{naraanja,orabys,cervon,gsig,chenyt,wenbzhao\}@amazon.com} \\
    \texttt{yc429@cam.ac.uk}, \texttt{\{tagyoung,jhuangz\}@amazon.com}
}

\begin{document}
\maketitle
\begin{abstract} 
Automatic melody-to-lyric generation is a task in which song lyrics are generated to go with a given melody.
It is of significant practical interest and more challenging than unconstrained lyric generation as the music imposes additional constraints onto the lyrics. The training data is limited as most songs are copyrighted, resulting in models that underfit the complicated cross-modal relationship between melody and lyrics. In this work, we propose a method for generating high-quality lyrics without training on any aligned melody-lyric data.  Specifically, we design a hierarchical lyric generation framework that first generates a song outline and second the complete lyrics. The framework enables disentanglement of training (based purely on text) from inference (melody-guided text generation) to circumvent the shortage of parallel data.

We leverage the segmentation and rhythm alignment between melody and lyrics to compile the given melody into decoding constraints as guidance during inference. The two-step hierarchical design also enables content control via the lyric outline, a much-desired feature for democratizing collaborative song creation.
Experimental results show that our model can generate high-quality lyrics that are more on-topic, singable, intelligible, and coherent than strong baselines, for example SongMASS \cite{sheng2021songmass}, a SOTA model trained on a parallel dataset, with a $24\%$ relative overall quality improvement based on human ratings. \footnote{Our code is available at \url{https://tinyurl.com/yhytyz9b}.}  
\end{abstract}

\section{Introduction}\label{sec:intro}
Music is ubiquitous and an indispensable part of humanity \cite{edensor2020national}. Self-serve songwriting has thus become an emerging task and has received interest by the AI community \cite{sheng2021songmass, tan2021tutorial, zhang2022relyme, guo-etal-2022-automatic}. However, the task of melody-to-lyric (M2L) generation, in which lyrics are generated based on a given melody, is underdeveloped due to two major challenges. 
First, there is a limited amount of melody-lyric aligned data. The process of collecting and annotating paired data is not only labor-intensive but also requires strong domain expertise and careful consideration of copyrighted source material. In previous work, either a small amount (usually a thousand) of melody-lyrics pairs is manually collected \cite{watanabe2018melody, lee2019icomposer}, or \citet{sheng2021songmass} use the recently publicized data \cite{yu2021conditional} in which the lyrics are pre-tokenized at the syllable level leading to less sensical subwords in the outputs. 

 \begin{figure}[]
  \centering
 \includegraphics[width=0.5\textwidth]{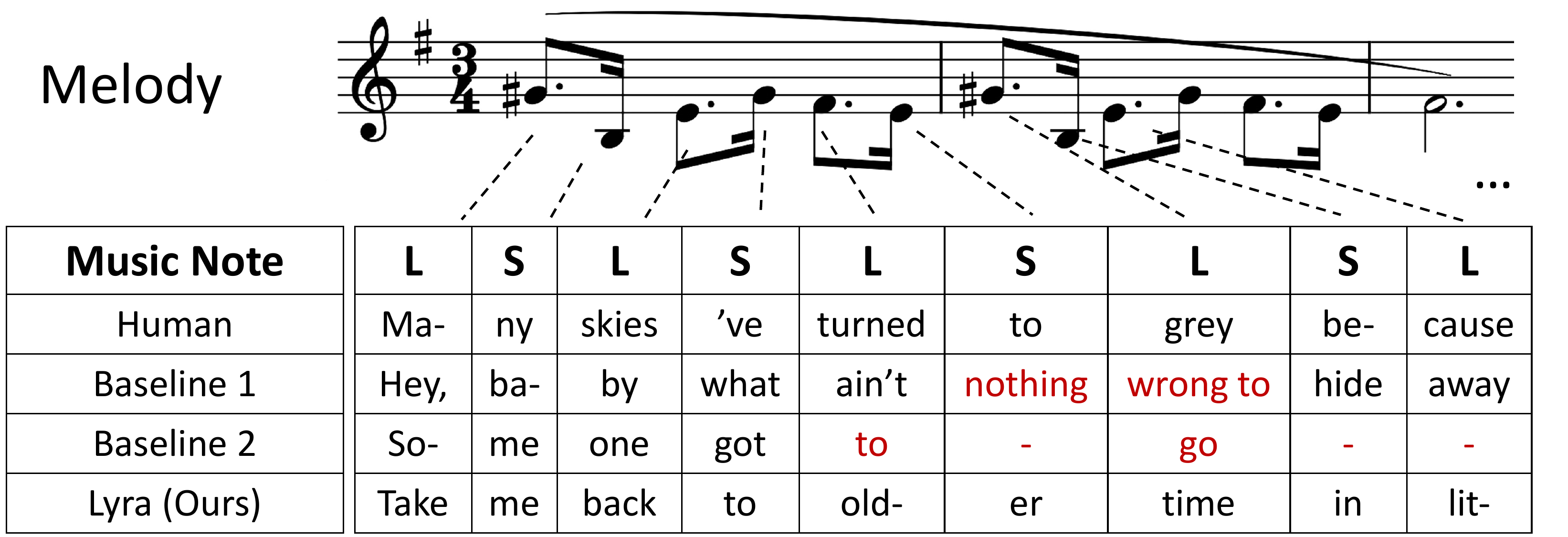}\vspace{-2mm}
  \caption{An example of the melody and the corresponding lyrics, where `L' denotes a music note with long duration and `S' stands for short. Our model \ModelName{} generates more coherently than the baselines. Besides, the rhythms of lyrics (i.e., accents and relaxations when spoken) generated by human and \ModelName{} align well with the flows of the melody. On the other hand, existing methods output lyrics that have low singability by either aligning multiple words with one single note (baseline 1) or vice versa (baseline 2) as highlighted in red.}
  \label{fig:example} 
  \vspace{-5mm}
\end{figure}

Another challenge lies in melody-to-lyric modeling. Compared to unimodal sequence-to-sequence tasks such as machine translation, the latent correlation between lyrics and melody is difficult to learn. For example,
\citet{watanabe2018melody, lee2019icomposer, chen2020melody,sheng2021songmass} apply RNNs, LSTMs, SeqGANs, or Transformers with melody embeddings and cross attention \cite{vaswani2017attention}, hoping to capture the melody-lyrics mapping. However, as shown in Figure \ref{fig:example}, these methods may generate less singable lyrics when they violate too often a superficial yet crucial alignment: one word in a lyric tends to match one music note in the melody \cite{nichols2009relationships}. In addition, their outputs are not fluent enough because they are neural models trained from scratch without leveraging large pre-trained language models (PTLMs). 

In this paper, we propose \textbf{\ModelName{}}, an unsupervised, hierarchical melody-conditioned LYRics generAtor that can generate high-quality lyrics with content control \textit{without training on melody-lyric data}. 
To circumvent the shortage of aligned data, \ModelName{} leverages PTLMs and disentangles training (pure text-based lyric generation) from inference (melody-guided lyric generation). This is motivated by the fact that plain text lyrics under open licenses are much more accessible \cite{tsaptsinos2017lyrics,lyrics-dataset,edmonds2021multi}, and prior music theories pointed out that the knowledge about music notes can be compiled into constraints to guide lyric generation. Specifically, \citet{dzhambazov2017knowledge} argue that it is the \textit{durations} of music notes, not the pitch values, that plays a significant role in melody-lyric correlation. 

As shown in Figure \ref{fig:example}, the segmentation of lyrics should match the segmentation of music phrases for breathability. \citet{oliveira2007tra,nichols2009relationships} also find that long (short) note durations tend to associate with (un)stressed syllables. However, existing lyric generators, even when equipped with state-of-the-art neural architectures and trained on melody-lyrics aligned data, still fail to capture these simple yet fundamental rules. 
In contrast, we show that through an inference-time decoding algorithm that considers two melody constraints (segment and rhythm) without training on melody-lyrics aligned data, \ModelName{} achieves better singability than the best data-driven baseline. Without losing flexibility, we also introduce a factor to control the strength of the constraints.

In addition, \ModelName{} adopts the hierarchical text generation framework (i.e., plan-and-write \citep{fan-etal-2019-strategies, yao2019plan}) that both helps with the coherence of the generation and improves the controllability of the model to accommodate user-specified topics or keywords. During training, the input-to-plan model learns to generate a plan of lyrics based on the input title and salient words, then the plan-to-lyrics model generates the complete lyrics. To fit in the characteristics of lyrics and melody, we also equip the plan-to-lyrics model with the ability to generate sentences with a predefined count of syllables through multi-task learning. 

Our contributions are summarized as follows: 
\begin{itemize}[topsep=0pt, itemsep=-2pt, leftmargin=*]
\item 
We design \ModelName{}, the first melody-constrained neural lyrics generator \textit{without training on parallel data}. Specifically, we propose a novel hierarchical framework that disentangles training from inference-time decoding, which is supported by music theories. Our method works with most PTLMs, including those black-box large language models (LLMs) when finetuning is replaced by in-context learning.

\item 
The hierarchical generation design of \ModelName{} enables content or topic control, a feature of practical interest but missing among existing works. 
\item Both automatic and human evaluations show that our unsupervised model \textbf{\ModelName{}} outperforms fully supervised baselines in terms of both text quality and musicality by a significant margin. \footnote{Examples of lyrics generated by the complete pipeline can be found \href{https://sites.google.com/view/lyricsgendemo}{in this demo page}.}
\end{itemize}

 \begin{figure*}[]
 \vspace{-2mm}
  \centering
 \includegraphics[width=1\textwidth]{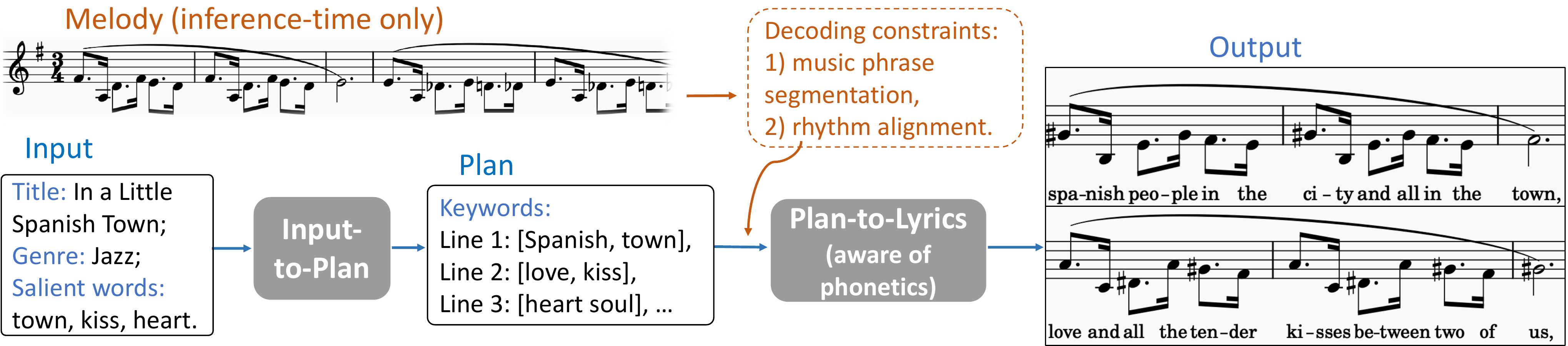}
  \vspace{-5mm}\caption{An overview of our approach that disentangles training from inference. {\color[HTML]{526EC0}Blue} represents components used during both training and inference, while {\color[HTML]{CD6607}brown} means inference only.  During training, our input-to-plan model learns to predict sentence-level plans (i.e., keywords) given the title, genre, and salient words as input. Then, the plan-to-lyrics model generates the lyrics while being aware of word phonetic information and syllable counts. At inference time, we compile the given melody into 1) music phrase segments and 2) rhythm constraints to guide the generation. \looseness=-1}
  \label{fig:overview}
  \vspace{-5mm}
\end{figure*}

\section{Background and Problem Setup}
\paragraph{Representation of Melody}
Melody is a succession of pitches in rhythm consisting of a sequence of music phrases, which can be further decomposed into timed music notes. Each music note is defined by two independent pivots: pitch values and durations. \textit{Pitch} represents the highness/lowness of a musical tone; \textit{duration} is the note's length of time.  Namely, melody $\mathcal{M}$ can be denoted by $\mathcal{M}=\{ p_1, p_2, ... p_M \}$, where each $p_i$ ($i \in {1,2, ...,M}$) is a music phrase. The music phrase can be further decomposed into timed music notes ($p_i = \{n_{i1}, n_{i2}, ... n_{iN_i}\}$), where each music note $n_{ij}$ ($j \in \{1,2, ...,N_i\}$) comes with a duration and is associated with or without a pitch value. When a music note comes without a pitch value, it is a rest that indicates the absence of a sound and usually aligns with no lyrics. 

\paragraph{Task Definition}
We follow the definition of ``unsupervised'' Machine Translation (MT) tasks \cite{lampleunsupervised, artetxe2019effective} which achieve cross-lingual translation by training on monolingual data only. 
In our case,  we achieve ``unsupervised'' melody-to-lyrics generation by training on text data only and do not require any parallel melody-lyrics aligned data for training.

\paragraph{Task Formulation}
We aim to generate lyrics that comply with both the provided topic and melody. The input topic is further decomposed into an intended title $T$ and a few salient words $S$ to be included in the generated lyrics (see Figure \ref{fig:overview} for an example input).  Following the settings of previous work \citep{chen2020melody, sheng2021songmass}, we assume that the input melody $\mathcal{M}$ is predefined and consists of $M$ music phrases ($\mathcal{M}=\{ p_1, p_2, ... p_M \}$), and each music phrase contains $N_i$ music notes ($p_i = \{n_{i1}, n_{i2}, ... n_{iN_i}\}$). 
The output is a piece of lyrics $\mathcal{L}$ that aligns with the music notes: $\mathcal{L} = \{ w_{11}, w_{12}, ..., w_{MN}\}$. Here, for $j \in \{1,2, ...,N_i\}$, $ w_{ij}$ is a word or a syllable of a word that aligns with the music note $n_{ij}$.

\section{Lyric Generation Model} 

We draw inspirations from recent generation models with content planning. These models are shown to achieve increased coherence and relevance over end-to-end generation frameworks in tasks such as story generation \cite{fan2018hierarchical, yao2019plan,yang2022re3}. Our lyrics generation model is similarly hierarchical as is shown in Figure \ref{fig:overview}. 
Specifically, we finetune two modules in our purely text-based pipeline: 1)
an input-to-plan generator that generates a keyword-based intermediate plan, and 2) a plan-to-lyrics generator which is aware of word phonetics and syllable counts.

\subsection{Input-to-Plan}
In real-world scenarios, users will likely have an intended topic (e.g., a title and a few keywords) to write about. We similarly extract a few salient words from the training lyric using the YAKE algorithm \cite{campos2020yake}, and feed them to our input-to-plan module to improve topic relevance. 
The input contains the song title, the music genre, and three salient words extracted from ground truth lyrics. Note that we chose 3 as a reasonable number for practical use cases, but our approach works for any arbitrary number of salient keywords. 

Our input-to-plan model is then trained to generate a line-by-line keyword plan of the song.
Considering that at inference time we may need different numbers of keywords for different expected output lengths, the number of planned keywords is flexible. Specifically, we follow the settings in \citet{tian2022zero} to include a placeholder (the $<$MASK$>$ token) in the input for every keyword to be generated in the plan. In this way, we have control over how many keywords we would like per line. We finetune BART-large \cite{lewis2019bart} 
as our input-to-plan generator with format control. \looseness=-1
\begin{table}[t]
\centering
\resizebox{\columnwidth}{!}{%
\begin{tabular}{|c|c|c|}
\hline
Model                      &  Output: Generated Lyric                \\ \hline
Naïve                &Cause the Christmas gift was for.             \\ \hline
\citet{chen2020melody}    &  Hey now that's what you ever. \\ \hline
\citet{sheng2021songmass} &  Believe you like taught me to.                        \\ \hline
Ours, Multi-task & Night and day my dreams come true.     \\ \hline
\end{tabular}
}
\caption{Examples of lyrics generated by different models with seven syllable counts as a constraint. Our model with multi-task auxiliary learning is the only system that successfully generates a complete line of lyrics with the desired number of syllables. On the other hand, the supervised models \citep{chen2020melody, sheng2021songmass} trained with melody-lyrics paired data still generate dangling or cropped lyrics.}
\label{table:dangling}    
\vspace{-5mm}
\end{table}

\subsection{Plan-to-Lyrics}\label{sec:p2l} Our plan-to-lyrics module takes in the planned keywords as input and generates the lyrics. This module encounters an added challenge: to match the music notes of a given melody at inference time, it should be capable of generating lyrics with a desired syllable count that aligns with the melody. If we naïvely force the generation to stop once it reaches the desired number of syllables, the outputs are usually cropped abruptly or dangling. For example, if the desired number of syllables is 7, a system unaware of this constraint might generate `Cause the Christmas gift was for' which is cropped and incomplete. Moreover, two recent lyric generators which are already trained on melody-to-lyrics aligned data also face the same issue (Table \ref{table:dangling}).

We hence propose to study an under-explored task of \textit{syllable planning}: generating a line of lyrics that 1) is a self-contained phrase and 2) has the desired number of syllables. To this end, we include both the intermediate plan and the desired syllable count as input. Additionally, we propose to equip the plan-to-lyrics module with the word phonetics information and the ability to count syllables. We then adopt multi-task auxiliary learning to incorporate the aforementioned external knowledge during training, as \citet{liebel2018auxiliary,guo2019autosem,poth2021pre,kung-etal-2021-efficient} have shown that related \textit{auxiliary tasks} help to boost the system performance on the \textit{target task}. Specifically, we study the collective effect of the following related tasks which could potentially benefit the model to learn the target task: \begin{itemize}[leftmargin=*]
\itemsep0.1em
\item T1: Plan to lyrics generation with syllable constraints (the target task)
\item T2: Syllable counting: given a sentence, count the number of syllables
\item T3: Plan to lyrics generation with granular syllable counting: in the output lyric of T1, append the syllable counts immediately after each word 
\item T4: Word to phoneme translation
\end{itemize}
\begin{table}[t]
\small
\centering
\begin{tabular}{ll}
\toprule
\multicolumn{1}{c}{\textbf{Task}} & \multicolumn{1}{c}{\textbf{Sample Data} ({\color[HTML]{656565}Input →} Output)}                                                                                                                                                \\ \midrule
T1                                  & \begin{tabular}[c]{@{}l@{}}{\color[HTML]{656565}Line 1: 8 syllables; Keywords: ...  →} \\ Line 1: Moon river wider than a mile; ....\end{tabular}                          \\ \midrule
T2                                  & {\color[HTML]{656565}Moon river wider than a mile →}  8                                                                                                                                             \\ \midrule
T3                                  & \begin{tabular}[c]{@{}l@{}}{\color[HTML]{656565}Line 1: 8 syllables; Keywords: ...  →} \\ Line 1: Moon (1) river (3) wider (5)\\\qquad than (6) a (7) mile (8); .... \end{tabular} \\ \midrule
T4                                  & \begin{tabular}[c]{@{}l@{}}{\color[HTML]{656565}Moon →}  MUWN; {\color[HTML]{656565}river →}  RIH\_VER;\\{\color[HTML]{656565}wider →} WAY\_DER; ... \end{tabular}                                                                                                             \\ \bottomrule
\end{tabular}
\caption{Sample data of the four proposed tasks to facilitate lyric generation with syllable planning.}
\label{table:sample-data} 
\vspace{-5mm}
\end{table}
We list the sample data for each task in Table \ref{table:sample-data}. We aggregate training samples from the above tasks, and finetune GPT-2 large \cite{radford2019language} on different combinations of the four tasks. We show our model's success rate on the target task in Table \ref{table:syllable-count} in Section \ref{subsec:num_syllable}. 

\section{Melody-Guided Inference}\label{sec:inf}
In this section, we discuss the procedure to compile a given melody into constraints to guide the decoding at inference time. We start with the most straightforward constraints introduced before: 1) segmentation alignment and 2) rhythm alignment. Note that both melody constraints can be updated without needing to retrain the model.

\subsection{Segment Alignment Constraints}
The segmentation of music phrases should align with the segmentation of lyrics \cite{watanabe2018melody}. Given a melody, we first parse the melody into music phrases, then compute the number of music notes within each music phrase. For example, the first music phrase in Figure \ref{fig:overview} consists of 13 music notes, which should be equal to the number of syllables in the corresponding lyric chunk.  Without losing generality, we also add variations to this constraint where multiple notes can correspond to one single syllable when we observe such variations in the gold lyrics.

\subsection{Rhythm Alignment Constraints}
According to \citeauthor{nichols2009relationships}, the stress-duration alignment rule hypothesizes that music rhythm should align with lyrics meter. Namely, shorter note durations are more likely to be associated with unstressed syllables. At inference time, we `translate' a music note to a stressed syllable (denoted by 1)  or an unstressed syllable (denoted by 0) by comparing its duration to the average note duration. For example, based on the note durations, the first music phrase in Figure \ref{fig:overview} is translated into alternating 1s and 0s, which will be used to guide the inference decoding.

\subsection{Phoneme-Constrained Decoding}
At each decoding step, we ask the plan-to-lyrics model to generate candidate complete words, instead of subwords, which is the default word piece unit for GPT-2 models. This enables us to retrieve the word phonemes from the CMU pronunciation dictionary \cite{weide1998carnegie} and identify the resulting syllable stresses. For example, since the phoneme of the word `Spanish' is `S PAE1 NIH0 SH', we can derive that it consists of 2 syllables that are stressed and unstressed. 

Next, we check if the candidate words satisfy the stress-duration alignment rule. Given a candidate word $w_i$ and the original logit $p(w_i)$ predicted by the plan-to-lyrics model, we introduce a factor $\alpha$ to control the strength:
\begin{equation}
    p'(w_i) =
    \begin{cases}
        p(w_i), &  \text{{\small if $w_i$ satisfies rhythm alignment,}}\\
        \alpha p(w_i), & \text{\small otherwise.}
    \end{cases}
\end{equation}
We can either impose a \textbf{hard constraint}, where we reject all those candidates that do not satisfy the rhythm rules ($\alpha$ = 0), or impose a \textbf{soft constraint}, where we would reduce their sampling probabilities ($0 < \alpha < 1$). Finally, we apply diverse beam search \cite{vijayakumar2016diverse} to promote the diversity of the generated sequences. 
\section{Experimental Setup}
In this section, we describe the train and test data, baseline models, and evaluation setup. The evaluation results are reported in Section \ref{sec:results}.

\subsection{Dataset}\label{sec:data}
\paragraph{Train data.} Our training data consists lyrics of 38,000 English songs and their corresponding genres such as Pop, Jazz, and Rock, which we processed from the \href{https://www.kaggle.com/datasets/mateibejan/multilingual-lyrics-for-genre-classification}{Genre Classification dataset} \cite{lyrics-dataset}. The phonetic information needed to construct the auxiliary tasks to facilitate the syllable count control is retrieved from the CMU pronunciation dictionary \cite{weide1998carnegie}.

 \paragraph{Automatic test data.} The testing setup is the complete diagram shown in Figure \ref{fig:overview}. Our input contains both the melody (represented in music notes and phases) and the title, topical, and genre information. Our test melodies come from from the lyric-melody aligned dataset \cite{yu2021conditional}. In total, we gathered 120 songs that do not appear in the training data. Because the provided lyrics are pre-tokenized at the syllable level (e.g. "a lit tle span ish town" instead of "a little spanish town"), we manually reconstructed them back into natural words when necessary. 

\paragraph{Two sets of human test data.} 
To facilitate human evaluation, we leverage an online \href{http://www.sinsy.jp/}{singing voice synthesizer} \cite{hono2021sinsy} to generate the sung audio clips. This synthesizer however requires files in the musicXML format that none of the existing datasets provide (including our automatic test data). Therefore, we manually collected 6 copyrighted popular songs and 14 non-copyrighted public songs from the \href{https://musescore.org/en}{musescore platform} that supports the musicXML format. 

The first set of \textit{pilot} eval data are these 20 pieces of melodies that come with ground truth lyrics. 
In addition, we composed a second, \textit{larger} set of 80 test data by pairing each existing melody with various other user inputs (titles and salient words). This second 
 eval set, which does not come with ground truth lyrics, is aimed at comparison among all the models.

\subsection{Baseline Models for Lyrics Generation}
We compare the following models. \textbf{1. SongMASS} \cite{sheng2021songmass} is a state-of-the-art (SOTA) song writing system which leverages masked sequence to sequence pre-training and attention based alignment for M2L generation. It requires melody-lyrics aligned training data while our model does not. \textbf{2. GPT-2 finetuned on lyrics} is a uni-modal, melody-unaware GPT-2 large model that is finetuned end-to-end (i.e., \textbf{title-to-lyrics}). In the automatic evaluation setting, we also compare an extra variation, \textbf{content-to-lyrics}, in which the input contains the title, salient words, and genre. 
These serve as ablations of the next model \textit{\ModelName{} w/o rhythm} to test the efficacy of our plan-and-write pipeline without inference-time constraints.
\textbf{3. \ModelName{} w/o rhythm} is our base model consisting of the input-to-plan and plan-to-lyrics modules with segmentation control, but without the rhythm alignment.
\textbf{4. \ModelName{} w/ soft/hard rhythm} is our multi-modal model with music segmentation and soft or hard rhythm constraints. For the soft constraints setting, the strength controlling hyperparameter $\alpha=0.01$. All models except SongMASS are finetuned on the same lyrics training data described in Section \ref{sec:data}.

\subsection{Automatic Evaluation Setup}
We automatically assess the generated lyrics on two aspects: the quality of text and music alignment. For \textbf{text quality}, we divide it into 3 subaspects: 1) \textbf{Topic Relevance}, measured by input salient word coverage ratio, and sentence- or corpus-level BLEU \cite{papineni2002bleu}; 2) \textbf{Diversity}, measured by distinct unigrams and bigrams \cite{li2015diversity}; 3) \textbf{Fluency}, measured by the perplexity computed using Huggingface's pretrained GPT-2. We also compute the ratio of cropped sentences among all sentences to assess how well they fit music phrase segments. For \textbf{music alignment}, we compute the percentage where the stress-duration rule holds.

\subsection{Human Evaluation Setup}
\paragraph{Turker Qualification} We used qualification tasks to recruit 120 qualified annotators who 1) have enough knowledge in song and lyric annotation, and 2) pay sufficient attention on the Mechanical Turk platform. The qualification consisted of two parts accordingly. 
First, to test the Turkers' domain knowledge, we created an annotation task consisting of the first verse from 5 different songs with gold labels. The 5 songs are carefully selected to avoid ambiguous cases, so that the quality can be clearly identified. We selected those whose scores have a high correlation with gold labels. Second, we adopted attention questions to rule out irresponsible workers. As is shown in the example questionnaire in Appendix \ref{appendix:survey}, we provided music sheets for each song in the middle of the questions. We asked all annotators the same question: ``Do you think the current location where you click to see the music sheet is ideal?''. Responsible answers include ``Yes'' or ``No'', and suggesting more ideal locations such as ``immediately below the audio clip and above all questions''.
We ruled out irresponsible Turkers who filled in geographical locations (such as country names) in the provided blank. 

\paragraph{Annotation Task} Our annotation is relative, meaning that annotators assess a group of songs generated from different systems with the same melody and title at once. We evaluated all baseline models except for GPT-2 finetuned (content-to-lyrics), as the two GPT-2 variations showed similar performance in automatic evaluation. We thus only included one due to resource constraints of the human study. Each piece of music was annotated by at least three workers, who were asked to evaluate the quality of the lyrics using a 1-5 Likert scale on six dimensions across musicality and text quality. For musicality, we asked them to rate \textbf{singability} (whether the melody's rhythm aligned well with the lyric's rhythm)  and \textbf{intelligibility} (whether the lyric content was easy to understand when listened to without looking at the lyrics).\footnote{The task was carefully designed so that intelligibility was asked before the workers read the lyrics. See Appendix \ref{appendix:survey} for more details.} For the lyric quality, we asked them to rate \textbf{coherence}, \textbf{creativeness}, and \textbf{in rhyme}. Finally, we asked annotators to rate how much they liked the song \textbf{overall}. 
A complete example of the survey can be found in Appendix \ref{appendix:survey}. 
The workers were paid \$16 per hour and the average inter-annotator agreement in terms of Pearson correlation was 0.47.

\begin{table}[t]
\centering
\small

\begin{tabular}{cccccc}
\toprule
\multicolumn{4}{c}{Task Name} & \multicolumn{2}{c}{Success Rate} \\
\midrule
\multicolumn{1}{c}{\begin{tabular}[c]{@{}c@{}}T1\\ \scriptsize{Lyrics}\end{tabular}} & \multicolumn{1}{c}{\begin{tabular}[c]{@{}c@{}}T2\\ \scriptsize{Count}\end{tabular}} & \multicolumn{1}{c}{\begin{tabular}[c]{@{}c@{}}T3\\\scriptsize{Granular}\end{tabular}} & \begin{tabular}[c]{@{}c@{}}T4\\\scriptsize{Phoneme}\end{tabular} & \multicolumn{1}{c}{\begin{tabular}[c]{@{}l@{}}Greedy\\ Decode\end{tabular}}  & {\begin{tabular}[c]{@{}l@{}}Sampling\\ Decode\end{tabular}}  \\ 
\midrule
\multicolumn{1}{c}{ $\checkmark$ }           & \multicolumn{1}{c}{}                                                                             & \multicolumn{1}{c}{}                                                                   &                                                                    & \multicolumn{1}{c}{23.14\%} & 19.87\% \\ 
\multicolumn{1}{c}{ $\checkmark$ }           & \multicolumn{1}{c}{$\checkmark$}                                                                             & \multicolumn{1}{c}{}                                                                   &  \multicolumn{1}{c}{}                                                                   & \multicolumn{1}{c}{50.14\%} & 44.64\% \\ 
\multicolumn{1}{c}{}            & \multicolumn{1}{c}{$\checkmark$}                                                                            & \multicolumn{1}{c}{$\checkmark$ }                                                                   &  \multicolumn{1}{c}{}                                                                   & \multicolumn{1}{c}{55.01\%} & 49.70\% \\ 
\multicolumn{1}{c}{ $\checkmark$ }           & \multicolumn{1}{c}{ $\checkmark$ }                                                                            & \multicolumn{1}{c}{$\checkmark$}                                                                   &  \multicolumn{1}{c}{}                                                                   & \multicolumn{1}{c}{\textbf{93.60\%}} & \textbf{89.13\% }\\ 
\multicolumn{1}{c}{ $\checkmark$ }           & \multicolumn{1}{c}{ $\checkmark$ }                                                                            & \multicolumn{1}{c}{ $\checkmark$ }                                                                  &  $\checkmark$                                                                   & \multicolumn{1}{c}{91.37\%} & 87.65\% \\ 
\bottomrule
\end{tabular}
\caption{Success rate for variants of our plan-to-lyrics model on generating sentences with the desired number of syllables.}
\label{table:syllable-count}
\vspace{-5mm}
\end{table}

\begin{table*}[t!]
\small
\centering
\begin{tabular}{|c|ccc|cc|c|cc|}
\hline
\rowcolor[HTML]{E7E6E6} 
\cellcolor[HTML]{E7E6E6}                             & \multicolumn{3}{c|}{\cellcolor[HTML]{E7E6E6}Content Control/Topic Relevance}                                                                                                                                                                                                              & \multicolumn{2}{c|}{\cellcolor[HTML]{E7E6E6}Diversity}             & \multicolumn{2}{c|}{\cellcolor[HTML]{E7E6E6}Fluency}     & Music                                                              \\ \cline{2-9} 
\rowcolor[HTML]{E7E6E6} 
\multirow{-2}{*}{\cellcolor[HTML]{E7E6E6}Model Name} & \multicolumn{1}{l|}{\cellcolor[HTML]{E7E6E6}\begin{tabular}[c]{@{}l@{}}Salient Word\\ Coverage↑\end{tabular}} & \multicolumn{1}{c|}{\cellcolor[HTML]{E7E6E6}\begin{tabular}[c]{@{}c@{}}Sent \\ Bleu↑\end{tabular}} & \begin{tabular}[c]{@{}c@{}}Corpus  \\ Bleu↑\end{tabular} & \multicolumn{1}{c|}{\cellcolor[HTML]{E7E6E6}Dist-1↑} & Dist-2↑        & PPL ↓       & \multicolumn{1}{c|}{\cellcolor[HTML]{E7E6E6}\begin{tabular}[c]{@{}c@{}}Cropped \\ Sentence↓\end{tabular}} & 
\multicolumn{1}{c|}{\cellcolor[HTML]{E7E6E6}\begin{tabular}[c]{@{}c@{}}Stress-\\Duration\end{tabular}}        \\ \hline
SongMASS {\scriptsize \cite{sheng2021songmass}}                                            & \multicolumn{1}{c|}{/}                                                                                       & \multicolumn{1}{c|}{0.045}                                                                        & 0.006                                                   & \multicolumn{1}{c|}{\textbf{0.17}}                  & \textbf{0.57} & 518         & \multicolumn{1}{c|}{34.51\%}                                                                            & 58.8\%          \\ \hline
GPT-2 (title-to-lyrics)                                    & \multicolumn{1}{c|}{/}                                                                                       & \multicolumn{1}{c|}{0.026}                                                                        & 0.020                                                   & \multicolumn{1}{c|}{0.09}                           & 0.31          & \textbf{82} & \multicolumn{1}{c|}{/}                                                                                  & 53.6\%          \\ \hline
GPT-2 (content-to-lyrics)                                    & \multicolumn{1}{c|}{83.3\%}                                                                                       & \multicolumn{1}{c|}{0.049}                                                                        & 0.027                                                   & \multicolumn{1}{c|}{0.10}                           & 0.42          & 87 & \multicolumn{1}{c|}{/}                                                                                  & 54.2\%          \\ \hline
\ModelName{} w/o rhythm                                      & \multicolumn{1}{c|}{\textbf{91.8\%}}                                                                         & \multicolumn{1}{c|}{\underline{0.074}}                                                                  & \underline{0.046}                                             & \multicolumn{1}{c|}{\underline{0.12}}                     & 0.45          & \underline{85}    & \multicolumn{1}{c|}{\textbf{3.65\%}}                                                                    & 63.1\%          \\ \hline
\ModelName{} w/   soft rhythm                                 & \multicolumn{1}{c|}{\underline{ 89.4\%}}                                                                            & \multicolumn{1}{c|}{\textbf{0.075}}                                                               & \textbf{0.047}                                          & \multicolumn{1}{c|}{0.11}                           & \underline{0.46}    & \underline{85}    & \multicolumn{1}{c|}{\underline{8.96\%}}                                                                       & \underline{68.4\%}    \\ \hline
\ModelName{} w/ hard rhythm                                 & \multicolumn{1}{c|}{88.7\%}                                                                                  & \multicolumn{1}{c|}{0.071}                                                                        & 0.042                                                   & \multicolumn{1}{c|}{\underline{0.12}}                     & 0.45          & 108         & \multicolumn{1}{c|}{10.26\%}                                                                            & \textbf{89.5\%} \\ \hline
\rowcolor[HTML]{E7E6E6} 
Ground   Truth                                       & \multicolumn{1}{c|}{\cellcolor[HTML]{E7E6E6}100\%}                                                           & \multicolumn{1}{c|}{\cellcolor[HTML]{E7E6E6}1.000}                                                & 1.000                                                   & \multicolumn{1}{c|}{\cellcolor[HTML]{E7E6E6}0.14}   & 0.58          & 93          & \multicolumn{1}{c|}{\cellcolor[HTML]{E7E6E6}3.92\%}                                                     & 73.3\%          \\ \hline
\end{tabular}\vspace{-2mm}
\caption{Automatic evaluation results. Human (ground truth) performance is highlighted in a grey background. Among all models, we highlight the best scores in boldface and underline the second best.}
  \vspace{-3mm}
  \label{table:auto_eval}  
\end{table*}

\section{Results}\label{sec:results}
\subsection{Generating a Sequence of Lyrics with the Desired Number of Syllables}\label{subsec:num_syllable}
Recall that in Section~\ref{sec:p2l}, we trained the plan-to-lyrics generator on multiple auxiliary tasks in order to equip it with the ability to generate a sentence with a pre-defined number of syllables. A sample output (boldfaced) can be found below: 
\textit{Line 1 (8 syllables): \textbf{Last Christmas I gave you my gift}; Line 2 (13 syllables): \textbf{It was some toys and some clothes that I said goodbye to}; Line 3 (11 syllables): \textbf{But someday the tree is grown with other memories}; Line 4 (7 syllables): \textbf{Santa can hear us singing};... }

\begin{figure*}[t]

\subfloat[\vspace{-1mm}Human evaluation results on the pilot test set with human as ground truth lyrics.]{%
  \vspace{-2mm}\includegraphics[clip,width=0.95\linewidth]{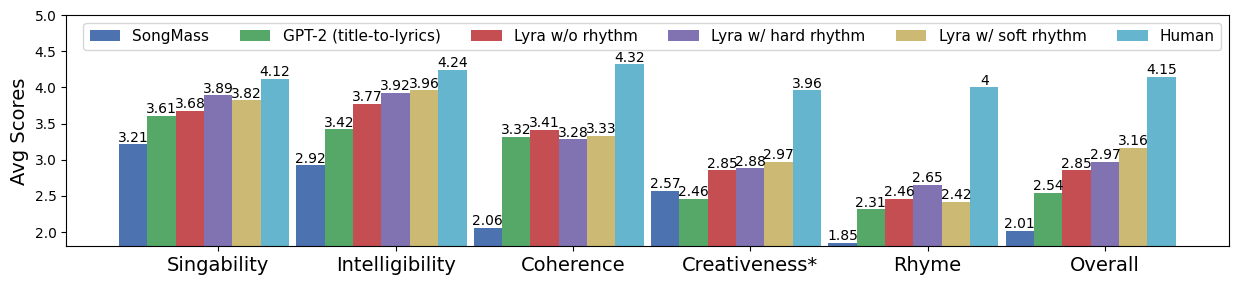}
\label{fig:human_eval1}   
}

\subfloat[Human evaluation results on the larger test set without ground truth lyrics.]{%
  \includegraphics[clip,width=0.95\linewidth]{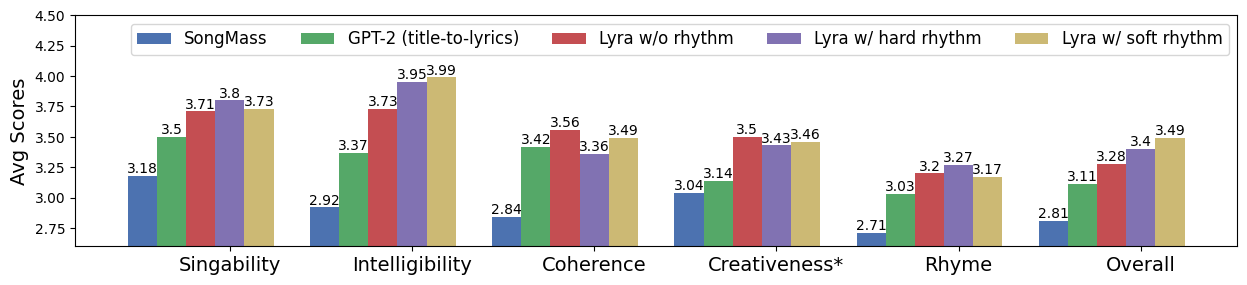}
  \label{fig:human_eval2}  
}
  \vspace{-1mm}
\caption{Average human Likert scores for two lyrics evaluation datasets on singability, intelligibility, coherence, creativity, rhyme, and overall quality. For each pair of systems in either study, we conduct paired t-test and observe statistical significance across all dimensions except creativeness (denoted by *).}
\vspace{-1.5em}

\end{figure*}
To test this feature, we compute the average success rate on a held-out set from the training data that contains 168 songs with 672 lines of lyrics. For each test sample, we compute its \textit{success} as a binary indicator where 1 indicates the output sequence contains exactly the same number of syllables as desired, and 0 for all other cases. We experimented with both greedy decoding and sampling, and found that BART \cite{lewis2019bart} could not learn
these multi-tasks as well as the GPT-2 family under the same settings. We hence report the best result of finetuning GPT-2 large \cite{radford2019language} in Table \ref{table:syllable-count}. 

The first row in Table \ref{table:syllable-count} shows that the model success rate is around 20\% without multi-task learning, which is far from ideal. By gradually training with auxiliary tasks such as syllable counts, the success rate increases, reaching over 90\% (rows 2, 3, 4). This shows the efficacy of multi-task auxiliary learning. We also notice that the phoneme translation task is not helpful for our goal (row 4), so we disregard the last task and only keep the remaining three tasks in our final implementation (row 3). 

\subsection{Automatic Evaluation Results}
We report the automatic evaluation results in Table \ref{table:auto_eval}. Our \ModelName{} models significantly outperform the baselines and generate the most on-topic and fluent lyrics. In addition, adding rhythm constraints to the base \ModelName{} noticeably increases the music alignment quality without sacrificing too much text quality. 
It is also noteworthy that humans do not consistently follow  stress-duration alignment, meaning that higher is not necessarily better for music alignment percentage. The comparisons between GPT-2 content-to-lyrics and \ModelName{} w/o rhythm support the hypothesis of the better topic control provided by our hierarchical architecture.

Since the baseline model SongMASS has no control over the content, it has lowest topic relevance scores. Moreover, although the SongMASS baseline seems to achieve the best diversity, it tends to produce non-sensical sentences that consist of a few gibberish words (e.g., `for hanwn to stay with him when, he got to faney he alone'), partially because its training data are pre-tokenized at the syllable level. Such degeneration is also reflected by the extremely high perplexity and cropped sentence ratio (CSR). Meanwhile, CSR is not applicable to both GPT-2 finetuned models because they are melody-unaware and generate lyrics freely without being forced to end at the end of each music segment.

\begin{figure*}[t!]\vspace{-1mm}
  \begin{subfigure}{0.5\textwidth}
    \centering
    \includegraphics[width=0.95\linewidth]{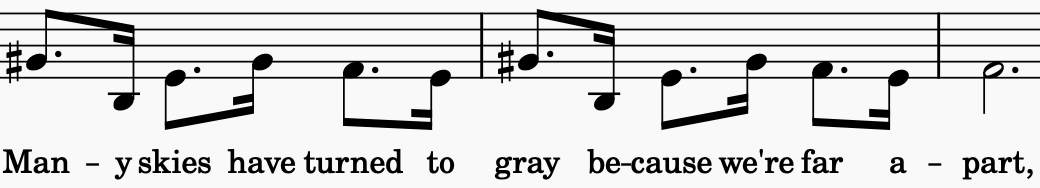}
    \caption{Human}
    \label{fig:1}
  \end{subfigure}%
  \begin{subfigure}{0.5\textwidth}
    \centering
    \includegraphics[width=0.95\linewidth]{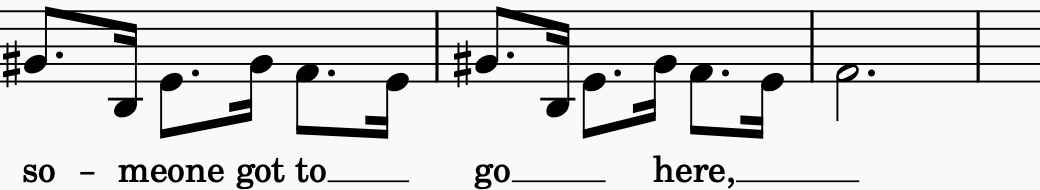}
    \caption{SongMASS}
    \label{fig:2}
  \end{subfigure}\medskip
  
  \begin{subfigure}{0.5\textwidth}
    \centering
    \includegraphics[width=0.95\linewidth]{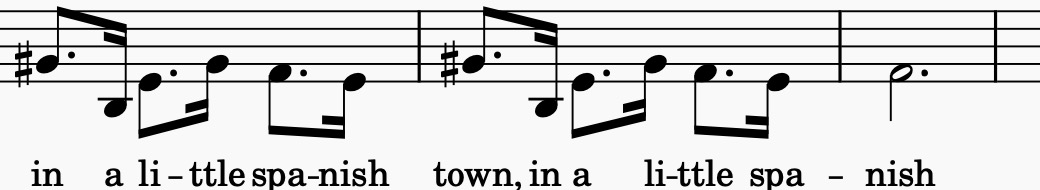}
    \caption{GPT-2 Finetuned on Title-to-Lyrics}
    \label{fig:3}
  \end{subfigure}
  \begin{subfigure}{0.5\textwidth}
    \centering
    \includegraphics[width=0.95\linewidth]{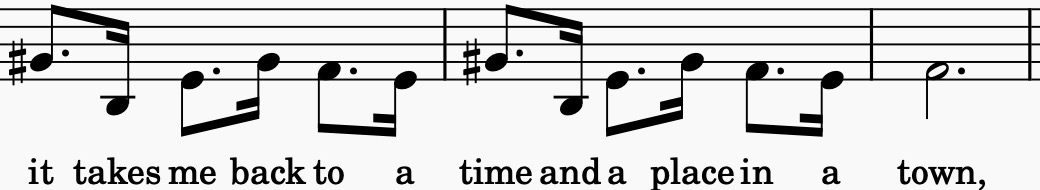}
    \caption{\ModelName{} w/o Rhythm}
    \label{fig:4}
  \end{subfigure}\medskip

  \begin{subfigure}{0.5\textwidth}
    \centering
    \includegraphics[width=0.95\linewidth]{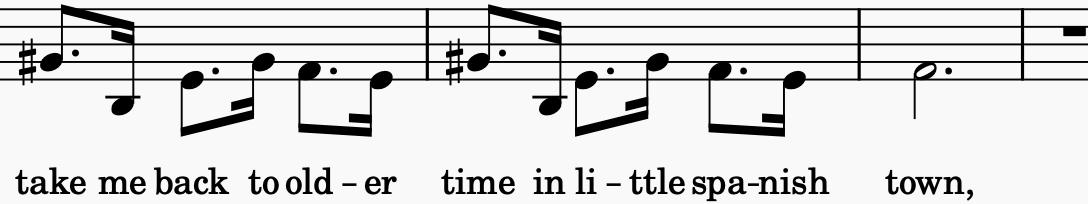}
    \caption{\ModelName{} w/ Soft Rhythm}
    \label{fig:5}
  \end{subfigure}
  \begin{subfigure}{0.5\textwidth}
    \centering
    \includegraphics[width=0.95\linewidth]{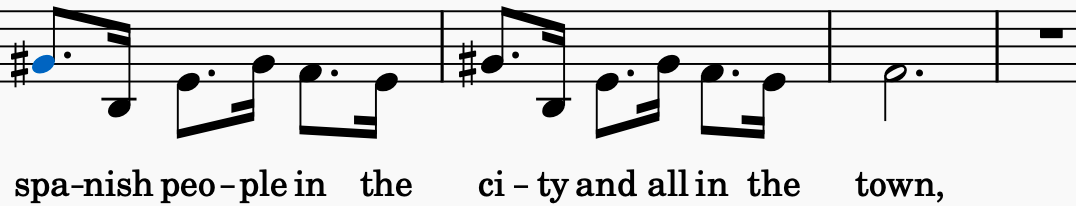}
    \caption{\ModelName{} w/ Hard Rhythm}
    \label{fig:6}
  \end{subfigure}\vspace{-2mm}
  \caption{Music sheets showing the lyric generated by different systems given the same piece of melody. \ModelName{} with soft and hard rhythm control are the only two models that can generate highly singable lyrics. The singing voices of the complete song can be found in this \href{https://sites.google.com/view/lyricsgendemo}{demo page}. }
  \label{fig:images}
\end{figure*}
\begin{table*}[t]
\small
\centering
\begin{tabular}{|L{7.5cm}|L{7.5cm}|}
\hline
\multicolumn{1}{|c|}{\textbf{a - Human} }                                                                                                                                                                                         & \multicolumn{1}{c|}{\textbf{b - SongMASS} \cite{sheng2021songmass} }                                                                                                                                                                               \\ \hline
Many skies have turned to gray because   we're far apart, Many moons have passed away and still she's in my heart, We made a promise and sealed it with a kiss, In a little Spanish town twas on a   night like this. & Someone got to go here, Forget that rest of my life, Everybody loves   somebody who i, In the middle of the night when the.                                                                                     \\ \hline
\multicolumn{1}{|c|}{\textbf{c - GPT-2 Finetuned on Title-to-Lyrics} }                                                                                                                                                                    & \multicolumn{1}{c|}{\textbf{d - \ModelName{} w/o Rhythm}}                                                                                                                                                                         \\ \hline
In a little Spanish town, In a little Spanish town, We'll make you feel good, We'll make you dance, In a little Spanish town, In a little Spanish town …                                                                                                                 & It takes me back to a time and a place in a \textit{town}, But if you do \textit{love} me  you will give me some \textit{kisses}, Where my \textit{heart} and \textit{soul} are not the same but   now, Your heart is in my \textit{heart} and not in a \textit{soul} i know. \\ \hline
\multicolumn{1}{|c|}{\textbf{e - \ModelName{} w/ Soft Rhythm} }                                                                                                                                                                           & \multicolumn{1}{c|}{\textbf{f - \ModelName{} w/ Hard Rhythm}  }                                                                                                                                                                   \\ \hline
Take me back to older time in little  \textit{Spanish} \textit{town}, And all the \textit{love} and all the \textit{kisses} that you gave me, I need   your \textit{heart} and your \textit{soul} and your love too, I need your \textit{heart} and \textit{soul} and I  need you back again. & \textit{Spanish} people in the city and all in the \textit{town}, \textit{Love} and all the tender \textit{kisses} between two of us, Is it my \textit{heart} or my \textit{soul} in you and me, In the   \textit{heart} and in the \textit{soul} and in the mind and then.           \\ \hline
\end{tabular}\vspace{-2mm}
\caption{An example of the generated lyrics with the same input - Title: In a little Spanish Town; Genre: Jazz; Salient words: town, kisses, heart. We highlight the generated keywords in italics. 
}
\label{table:moon-river}
\vspace{-5mm}
\end{table*}

\subsection{Human Evaluation Results}
The results on both evaluation sets are shown in Figures \ref{fig:human_eval1} and \ref{fig:human_eval2}. Clearly, human-written lyrics greatly outperform all models. For both evaluation sets, we notice the relative rankings of the models remain the same across all metrics except creativeness. This observation is mirrored by paired t-tests where we find that the best machine model differentiates from the second best machine model with statistical significance (p-value < 0.05) for all aspects except creativeness. Both indicate the reliability of our collected results in singability, intelligibility, coherence, rhyme, and overall quality.

\ModelName{} with hard or soft rhythm constraint are the best models in terms of singability, intelligibility, rhyme, and overall quality, which demonstrates the efficacy of our plan-and-write pipeline with melody alignment. We regard \ModelName{} with soft rhythm as our best model since it has highest overall quality.  The addition of soft rhythm alignment leads to further improvements in musicality and overall quality, with only a little sacrifice in coherence compared to GPT-2 (title-to lyrics). On the other hand, imposing hard rhythm constraints sacrifices the coherence and intelligibility of lyrics.

Surprisingly, SongMASS performs even worse than the finetuned GPT-2 baseline in terms of musicality. Upon further inspection, we posit that SongMASS too often deviates from common singing habits: it either assigns two or more syllables to one music note, or matches one syllable with three or more consecutive music notes.

\subsection{Qualitative Analysis}
We conduct a case study on an example set of generated lyrics to better understand the advantages of our model over the baselines. In this example, all models generate lyrics given the same title, genre, and salient words, as well as the melody of the original song.
We show the music sheet of the first generated segment in Figure \ref{fig:images} and the complete generated lyrics in Table \ref{table:moon-river}. We also provide the song clips with synthesized singing voices and more examples in this \href{https://sites.google.com/view/lyricsgendemo}{demo website}.

\paragraph{Musicality.} The melody-lyric alignment in Figure \ref{fig:images} is representative in depicting the pros and cons of the compared models. Although SongMASS is supervised on parallel data, it still often assigns too many music notes to one single syllable, which reduces singability and intelligibility. The GPT-2 title-to-lyrics model is not aware of the melody and thus fails to match the segmentation of music phrase with the generated lyrics. \ModelName{} w/o rhythm successfully matches the segments, yet stressed and long vowels such as in the words `takes' and `place' are wrongly mapped to short notes. Humans, as well as our models with both soft and hard rhythm alignment, produce singable lyrics.

\paragraph{Text quality.} As shown in Table \ref{table:moon-river}, SongMASS tends to generate simple and incoherent lyrics because it is trained from scratch. The GPT-2 title-to-lyrics model generates coherently and fluently, but is sometimes prone to repetition. All three variations of \ModelName{} benefit from the hierarchical planning stage and generate coherent and more informative lyrics.  However, there is always a \textbf{trade-off between musicality and text quality}. Imposing hard rhythm constraints could sometimes sacrifice coherence and creativity and thus hurt the overall quality of lyrics.

\section{Related Work}

\subsection{Melody Constrained Lyrics Generation}
\paragraph{End-to-End Models.} Most existing works on M2L generation are purely data-driven and suffer from a lack of aligned data. For example,\citet{watanabe2018melody, lee2019icomposer, chen2020melody} naively apply SeqGAN \cite{yu2017seqgan} or RNNs to sentence-level M2L generation. The data collection process is hard to automate and leads to manual collection of only small amounts of samples. Recently, \citet{sheng2021songmass} propose SongMASS by training two separate transformer-based models for lyric or melody with cross attention. 
To the best of our knowledge, our model \ModelName{} is the first M2L generator that does not require any paired cross-modal data, and is trained on a readily available uni-modal lyrics dataset.

\paragraph{Integrating External Knowledge.} \citet{oliveira2007tra,oliveira2015tra} apply rule-based text generation methods with predefined templates and databases for Portuguese. \citet{ma2021ai} use syllable alignments as reward for the lyric generator. However, it only estimates the expected number of syllables from the melody. We not only provide a more efficient solution to syllable planning, but also go one step further to incorporate the melody's rhythm patterns by following music theories \cite{nichols2009relationships, dzhambazov2017knowledge}. Concurrently, \citet{xue2021deeprapper,guo-etal-2022-automatic} partially share similar ideas with ours and leverage the sound to generate Chinese raps or translate lyrics via alignment constraints. Nevertheless, the phonetics of Chinese characters are very different from English words, and rap generation or translation is unlike M2L generation. 

\subsection{NLG with Hierarchical Planning}
Hierarchical generation frameworks are shown to improve consistency over sequence-to-sequence frameworks in other creative writing tasks such as story generation \cite{fan2018hierarchical, yao2019plan}. Recently, a similar planning-based scheme is adopted to poetry generation \cite{tian2022zero} to circumvent the lack of poetry data. We similarly equip \ModelName{} with the ability to comply with a provided topic via such content planning. 

\subsection{Studies on Melody-Lyrics Correlation}Music information researchers have found that it is the duration of music notes, not the pitch values that a play significant role in melody-lyric alignment \cite{nichols2009relationships, dzhambazov2017knowledge}. Most intuitively, one music note should not align with two or more syllables, and the segmentation of lyrics should match the segmentation of music phrases for singability and breathability \cite{watanabe2018melody}. In addition, \citet{nichols2009relationships} find out that there is a correlation between syllable stresses and note durations for better singing rhythm. Despite the intuitiveness of the aforementioned alignments, our experiments show that existing lyric generators which are already trained on melody-lyrics aligned data still tend to ignore these fundamental rules and generate songs with less singability. 
\section{Conclusion and Future Work}

Our work explores the potential of lyrics generation without training on lyrics-melody aligned data.
To this end, we design a hierarchical plan-and-write framework that disentangles training 
from inference
. At inference time, we compile the given melody into music phrase segments and rhythm constraints.  Evaluation results show that our model can generate high-quality lyrics that significantly outperform the baselines. Future directions include investigating more ways to compile melody into constraints such as the beat, tone or pitch variations, and generating longer sequences of lyrics with song structures such as verse, chorus, and bridge. Future works may also take into account different factors in relation to the melody such as mood and theme.

\section*{Acknowledgements}
The authors would like to thank the anonymous reviewers for the helpful comments.

\section*{Limitations}
We discuss the limitations of our work. First of all, our model \ModelName{} is build upon pre-trained language models (PTLM) including Bart \cite{lewis2019bart} and GPT-2 \cite{radford2019language}. Although our method is much more data friendly than previous methods in that it does not require training on melody-lyric aligned data, our pipeline may not apply to low-resource languages which do not have PTLMs. Second, our current adoption of melody constraints is still simple and based on a strong assumption of syllable stress and note duration. We encourage future investigation about other alignments such as the tone or pitch variations. Lastly, although we already have the music genre as an input feature, it remains an open question how to analyze or evaluate the generated lyrics with respect to a specific music genre.

\section*{Ethics Statement}
It is known that the generated results by PTLMs could capture the bias reflected in the training data \cite{sheng2019woman,wallace2019universal}. 
Our models may potentially generate offensive content for certain groups or individuals.  We suggest to carefully examine the potential biases before deploying the models to real-world applications.

\bibliography{anthology,custom}
\bibliographystyle{acl_natbib}

\clearpage
\appendix

\section{Survey Form Used In Human Evaluation}\label{appendix:survey}

\begin{figure*}[]
 \vspace{-2mm}
  \centering
 \includegraphics[width=1.0\textwidth]{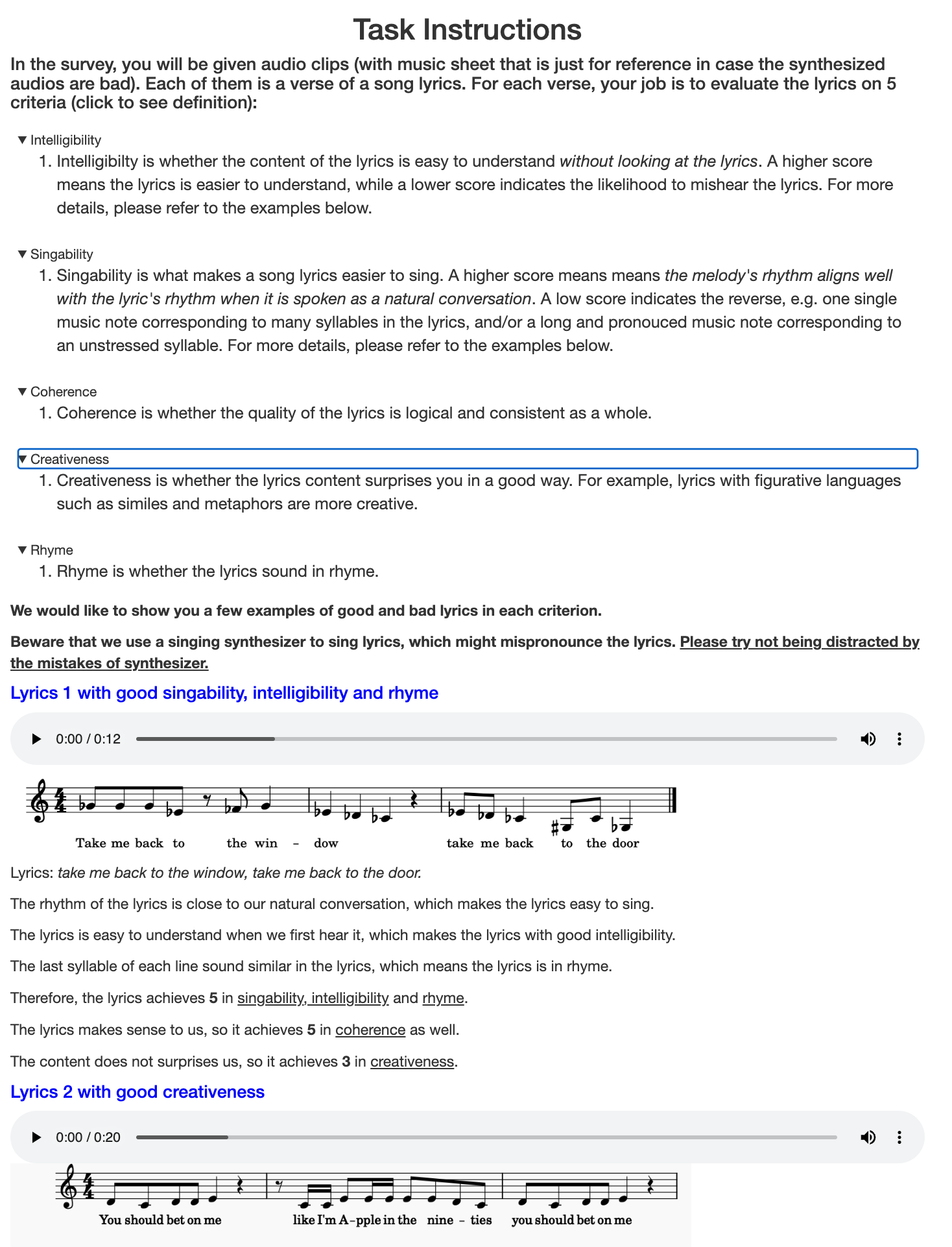}
  \vspace{-5mm}\caption{Task Instruction Page 1}
  \label{fig:Instruction1}
  \vspace{-5mm}
\end{figure*}

\begin{figure*}[]
 \vspace{-2mm}
  \centering
 \includegraphics[width=1.0\textwidth]{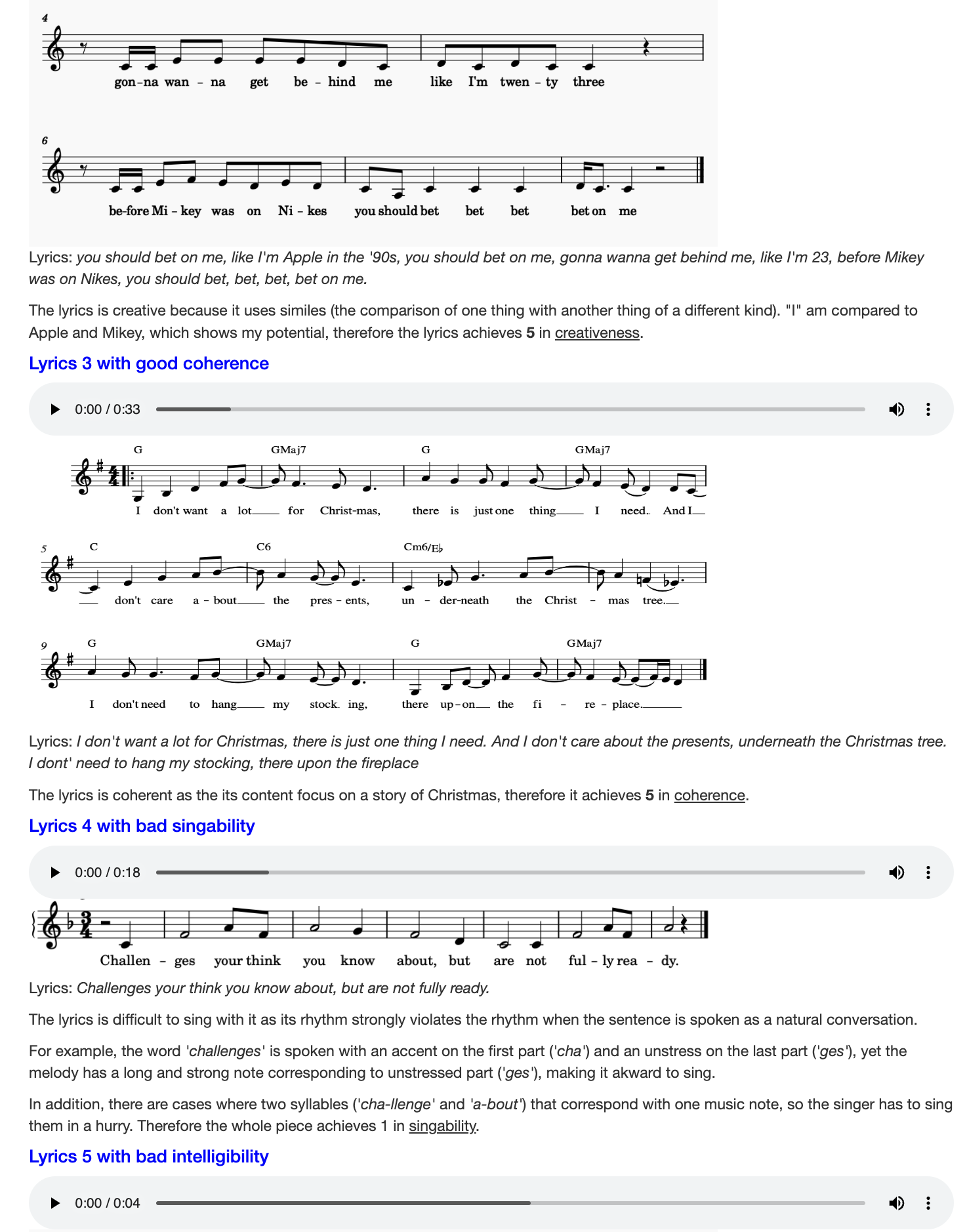}
  \vspace{-5mm}\caption{Task Instruction Page 2}
  \label{fig:Instruction2}
  \vspace{-5mm}
\end{figure*}

\begin{figure*}[]
 \vspace{-2mm}
  \centering
 \includegraphics[width=1.0\textwidth]{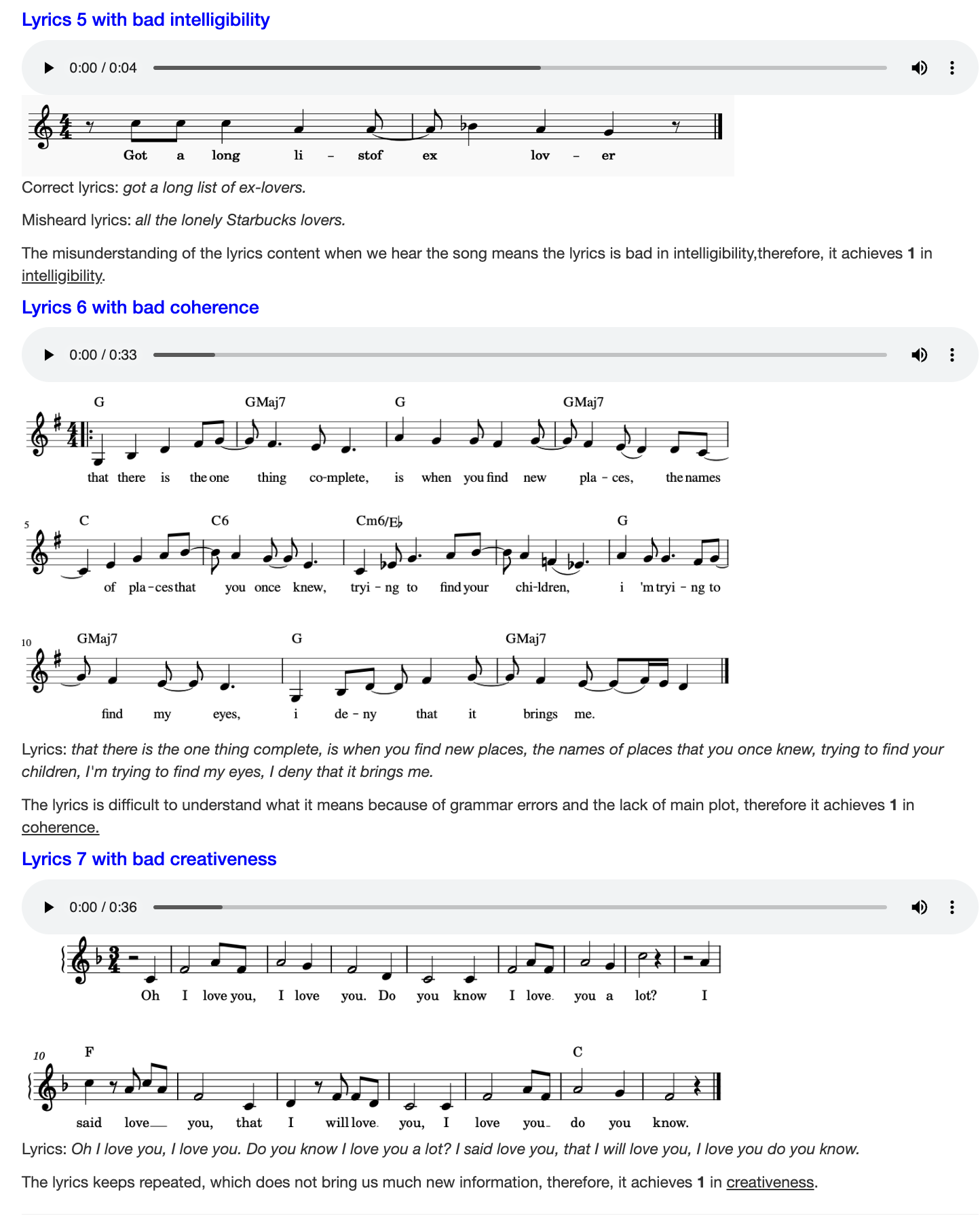}
  \vspace{-5mm}\caption{Task Instruction Page 3}
  \label{fig:Instruction3}
  \vspace{-5mm}
\end{figure*}

\begin{figure*}[]
 \vspace{-2mm}
  \centering
 \includegraphics[width=1.0\textwidth]{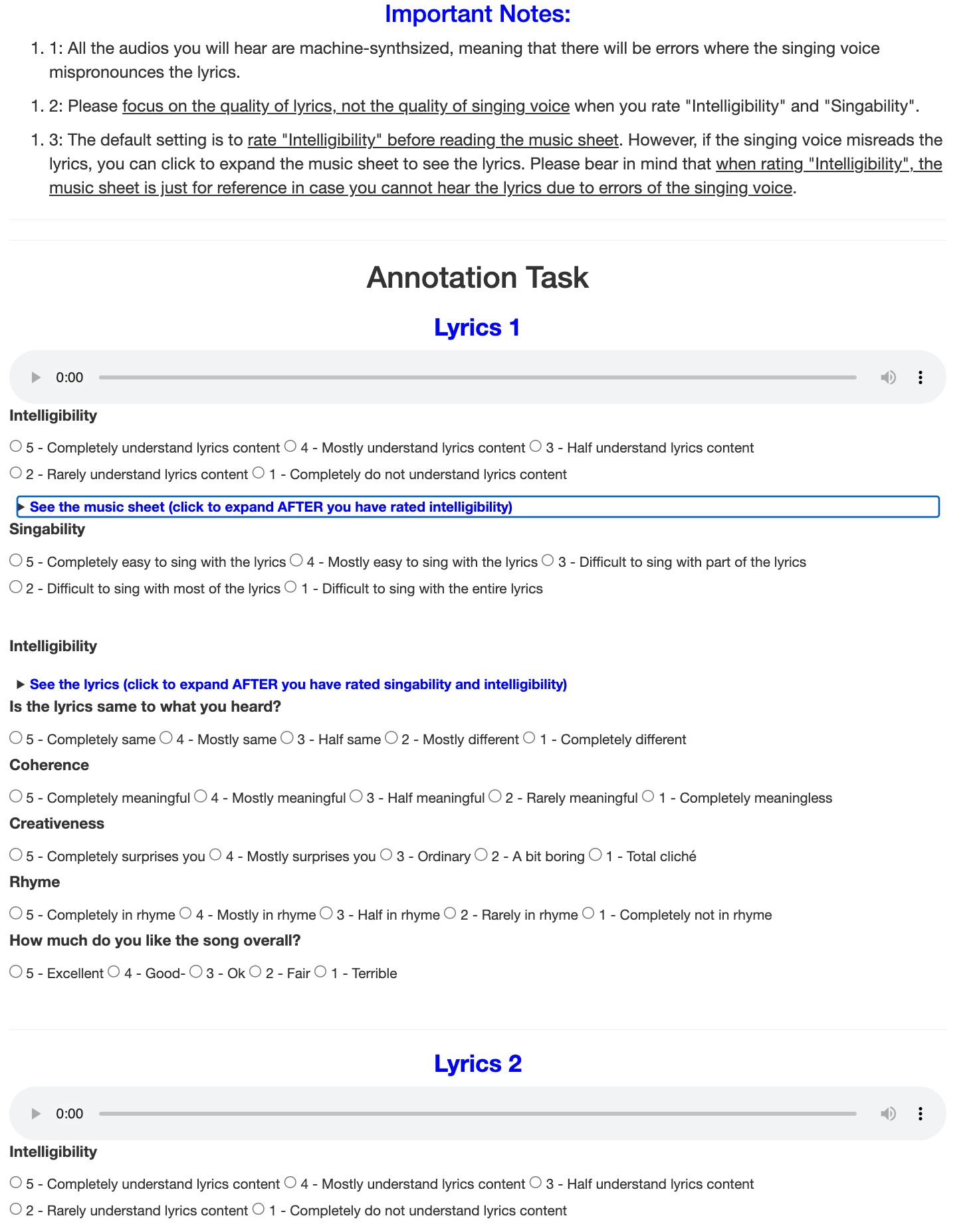}
  \vspace{-5mm}\caption{Annotation Task Page 1. We explicitly asked the annotators to rate Intelligibility twice, before and after they saw the generated lyrics and provided musicality scores. See explanations in Appendix \ref{appendix:survey}.}
  \label{fig:Task1}
  \vspace{-5mm}
\end{figure*}

\begin{figure*}[]
 \vspace{-2mm}
  \centering
 \includegraphics[width=1.0\textwidth]{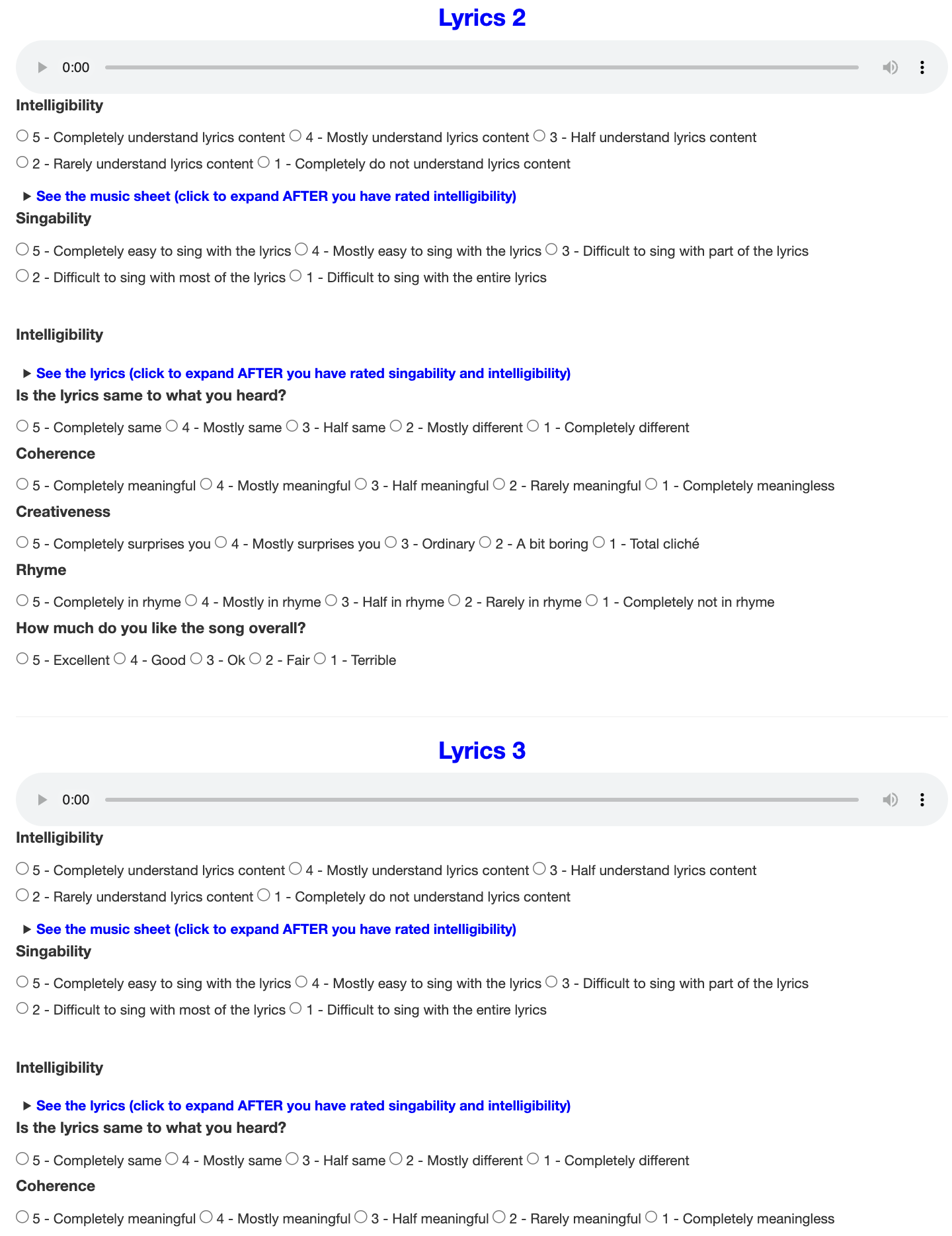}
  \vspace{-5mm}\caption{Annotation Task Page 2. We explicitly asked the annotators to rate Intelligibility twice, before and after they saw the generated lyrics and provided musicality scores.}
  \label{fig:Task2}
  \vspace{-5mm}
\end{figure*}

We show the original survey with the evaluation instructions and the annotation task in Figures \ref{fig:Instruction1} through \ref{fig:Task2}. Figure \ref{fig:Instruction1}, Figure \ref{fig:Instruction2}, and Figure \ref{fig:Instruction3} provide task instructions, including the definition of each metric (Intelligibility, Singability, Coherence, Creativeness, and Rhyme), and examples of good and bad lyrics in each criterion. Figures \ref{fig:Task1} and \ref{fig:Task2} showcase the actual annotation task. 

In the the actual annotation tasks, we noticed that annotators tended to adjust their rating to \textit{Intelligibility} (whether the content of the lyrics was easy to understand \textit{without looking at the lyrics}) after they were prompted to see the lyrics texts. We hence explicitly asked them to rate \textit{Intelligibility} twice, both before and after they saw the generated lyrics and music scores. Annotators must not modify their ratings to the first question after they saw the lyric texts, but could still use the second question to adjust their scores if needed. Such a mechanism helped us reduce the noise introduced by the presentation of lyric texts and music sheets. Namely, we asked the same questions twice, but only took into account the first intelligibility ratings when we computed the results.

\end{document}